\pdfoutput=1

\documentclass[11pt]{article}

\usepackage[preprint]{acl}

\usepackage{times}
\usepackage{latexsym}
\usepackage{amsmath}
\usepackage{multirow}
\usepackage{booktabs}
\usepackage{bbding}

\usepackage[T1]{fontenc}

\usepackage[utf8]{inputenc}

\usepackage{microtype}

\usepackage{inconsolata}

\usepackage{graphicx}

\title{On the Robustness of Document-Level Relation Extraction Models\\to Entity Name Variations}

\author{
 \textbf{Shiao Meng\textsuperscript{$\mathrm{1}$},}\quad
 \textbf{Xuming Hu\textsuperscript{$\mathrm{2}$},}\quad
 \textbf{Aiwei Liu\textsuperscript{$\mathrm{1}$},}\quad
 \textbf{Fukun Ma\textsuperscript{$\mathrm{1}$},}
\\
 \textbf{Yawen Yang\textsuperscript{$\mathrm{1}$},}\quad
 \textbf{Shuang Li\textsuperscript{$\mathrm{3}$},}\quad
 \textbf{Lijie Wen\textsuperscript{$\mathrm{1}$}\footnotemark[1]}
\\
 \textsuperscript{$\mathrm{1}$}School of Software, Tsinghua University
\\
 \textsuperscript{$\mathrm{2}$}AI Thrust, The Hong Kong University of Science and Technology (Guangzhou)
\\
 \textsuperscript{$\mathrm{3}$}Tencent Inc.
\\
 \texttt{msa21@mails.tsinghua.edu.cn, wenlj@tsinghua.edu.cn}
}

\begin{document}

\maketitle
\renewcommand{\thefootnote}{\fnsymbol{footnote}}
\footnotetext[1]{Corresponding author.}
\renewcommand{\thefootnote}{\arabic{footnote}}

\begin{abstract}
Driven by the demand for cross-sentence and large-scale relation extraction, document-level relation extraction (DocRE) has attracted increasing research interest. Despite the continuous improvement in performance, we find that existing DocRE models which initially perform well may make more mistakes when merely changing the entity names in the document, hindering the generalization to novel entity names. To this end, we systematically investigate the robustness of DocRE models to entity name variations in this work. We first propose a principled pipeline to generate entity-renamed documents by replacing the original entity names with names from Wikidata. By applying the pipeline to DocRED and Re-DocRED datasets, we construct two novel benchmarks named Env-DocRED and Env-Re-DocRED for robustness evaluation. Experimental results show that both three representative DocRE models and two in-context learned large language models consistently lack sufficient robustness to entity name variations, particularly on cross-sentence relation instances and documents with more entities. Finally, we propose an entity variation robust training method which not only improves the robustness of DocRE models but also enhances their understanding and reasoning capabilities. We further verify that the basic idea of this method can be effectively transferred to in-context learning for DocRE as well.\footnote{The data and code are available at \url{https://github.com/THU-BPM/Env-DocRE}.}
\end{abstract}

\section{Introduction}

\begin{figure}[t]
    \centering
    \includegraphics[width=0.9\linewidth]{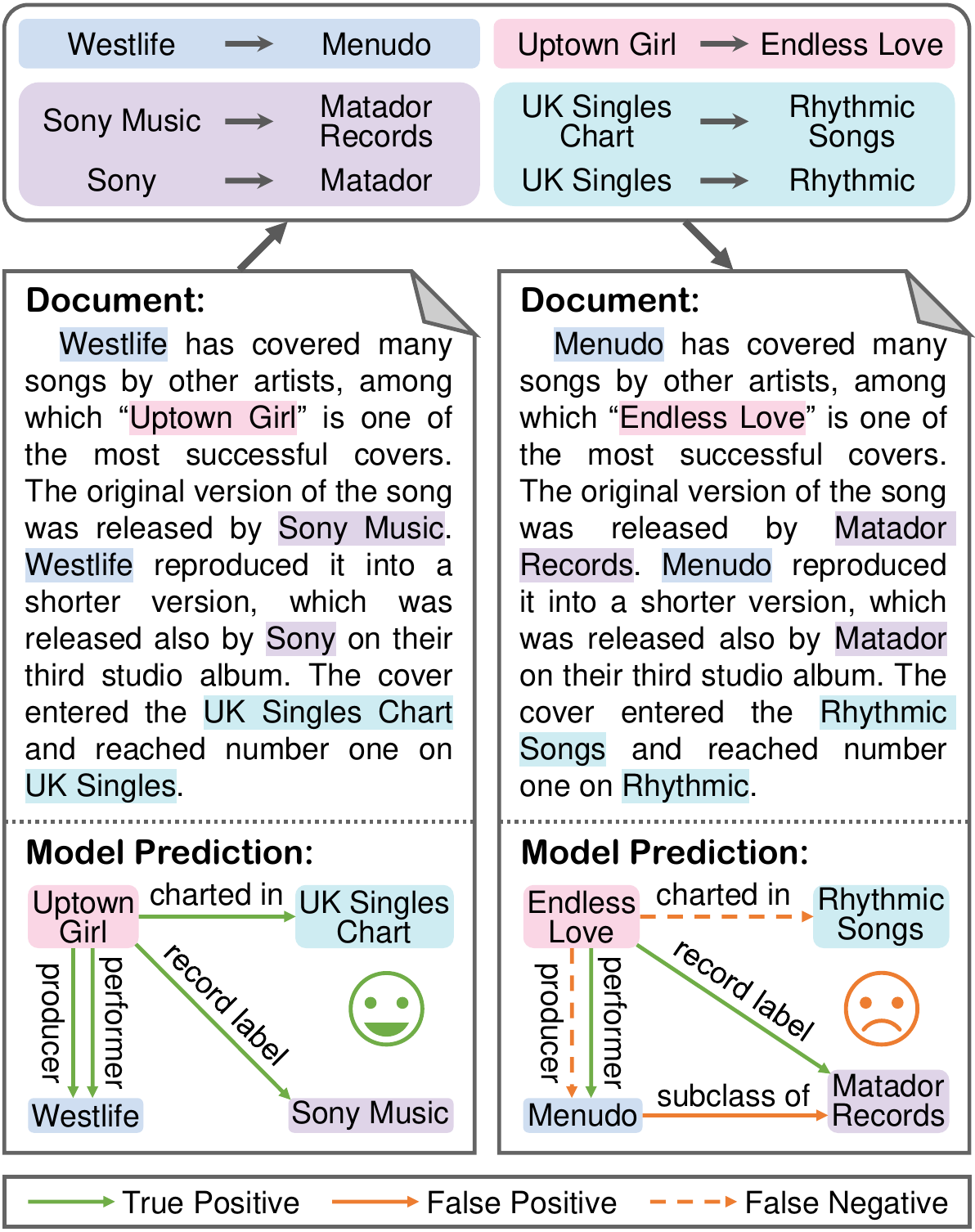}
    \caption{An illustration of how minor changes in entity names mislead the DocRE model to wrong predictions.}
    \label{fig:illustrative_example}
\end{figure}

The demand for cross-sentence and large-scale relation extraction has led to a surge of research interest in document-level relation extraction (DocRE), which aims to identify the relations between each pair of entities within a document based on the document context \citep{yao-etal-2019-docred}. While covering more realistic scenarios than its sentence-level counterpart \citep{hu2023selflre}, DocRE also brings new challenges, requiring a comprehensive modeling of interactions among different mentions of an entity, different entities and different entity pairs.

Recently, a series of DocRE studies propose various novel models and methods, continuously improving the performance on several DocRE benchmarks \citep{tan-etal-2022-document, zhou-lee-2022-ncrl, xiao-etal-2022-sais, sun-etal-2023-uncertainty}. However, we observe that existing DocRE models may produce more erroneous predictions when we merely change the entity names in a test document. As illustrated in Figure~\ref{fig:illustrative_example}, a well-trained DocRE model correctly extracts all four relation instances from the original document. However, after replacing the entity names in the document with a new set of names of the same entity types (e.g., change the song name \texttt{Uptown} \texttt{Girl} into another song name \texttt{Endless} \texttt{Love}), the model starts making mistakes, including both false positive and false negative predictions. This indicates that existing DocRE models may overly rely on entity information for extraction and lack robustness. Considering the vast and diverse space of entity names in real-world scenarios, which also expands constantly with numerous novel entity names, the poor robustness and generalization further impedes the reliable application of DocRE models.

As a result, we systematically study the robustness of DocRE models to entity name variations in this work. To audit the robustness of existing DocRE models, we first propose a general pipeline to automatically generate perturbed test documents with changed entity names. Building such a pipeline is non-trivial for three reasons: (1) The entity types are constrained by relation types in a fine-grained manner. For instance, the tail entity of relation \texttt{record} \texttt{label} in Figure~\ref{fig:illustrative_example} must be a record label. Therefore, the new entity name should not alter the original fine-grained entity type, otherwise the relation labels may no longer hold. (2) For an entity mentioned multiple times under different names, each alias should be replaced with a distinct name to exclude the interference caused by different coreference structures, like \texttt{Sony} \texttt{Music} $\Rightarrow$ \texttt{Matador} \texttt{Records} and \texttt{Sony} $\Rightarrow$ \texttt{Matador} in Figure~\ref{fig:illustrative_example}. (3) The introduced entity names should be of high quality and come from a wide range of sources. We strictly adhere to these three principles and design a four-stage pipeline based on Wikidata, which retrieves valid items from Wikidata for entity name substitution.

We further apply the proposed pipeline to DocRED \citep{yao-etal-2019-docred} and Re-DocRED \citep{tan-etal-2022-revisiting}, due to both being the largest and most widely used DocRE datasets, to create two novel benchmarks, named Env-DocRED and Env-Re-DocRED, for evaluating the robustness of DocRE models to entity name variations\footnote{Our proposed pipeline can also be applied or adapted to other DocRE datasets, which we discuss in detail in Section~\ref{sec:limitations_and_future_directions}.}. By conducting extensive experiments on both original and newly constructed benchmarks, we thoroughly evaluate the robustness of three representative DocRE models and two in-context learned large language models (LLMs). The results show that the performance of all evaluated models drops significantly on Env-DocRED and Env-Re-DocRED (e.g., the best model's F1 drops from 79.3\% on Re-DocRED to 57.0\% on Env-Re-DocRED), revealing the poor robustness to entity name variations. Further analyses reveal that the performance decline mainly lies in the increase of false negative predictions, and is more pronounced on cross-sentence relation instances and documents with more entities. We also analyze the reasons for performance drop by examining the loss of entity knowledge and name clues under entity name variations.

Finally, to improve the robustness of DocRE models to entity name variations, we propose an \textbf{E}ntity \textbf{V}ariation \textbf{R}obust \textbf{T}raining method (EVRT) which is based on data augmentation and consistency regularization. For each training document, we generate a perturbed document by entity renaming. Then, in addition to the classification loss for entity pairs in the original document, our method introduces three extra objectives, which respectively penalize the classification errors for entity pairs in the perturbed document, the inconsistency between representations, and inconsistency between predictions of original and corresponding perturbed entity pairs. Experimental results demonstrate that EVRT not only improves the robustness of existing DocRE models but also enhances their understanding and reasoning capabilities. Besides, we transfer the idea of EVRT to in-context learning of LLMs and propose a simple prompt optimization strategy, which effectively enhances the robustness of in-context learning for DocRE.
\section{Related Work}

\paragraph{Document-Level Relation Extraction.} Driven by the demand for cross-sentence and large-scale relation extraction, research on relation extraction has expanded from sentence level to document level \citep{quirk-poon-2017-distant, yao-etal-2019-docred}. Recently document-level relation extraction has attracted increasing research interest, with new models emerging constantly. Based on the way of modeling relational information from the context, existing studies can be categorized into graph-based and sequence-based approaches. The former typically abstract the document by graph structures and perform inference with graph neural networks \citep{zeng-etal-2020-double, zhang-etal-2021-docunet, wei2022sagdre, lu-etal-2023-anaphor}, while the latter encode the long-distance contextual dependencies with transformer-only architectures \citep{zhou-etal-2021-atlop, xie-etal-2022-eider, zhang-etal-2022-towards-better, meng-etal-2023-rapl, ma-etal-2023-dreeam, 10255305}.

\begin{figure*}[t]
    \centering
    \includegraphics[width=\linewidth]{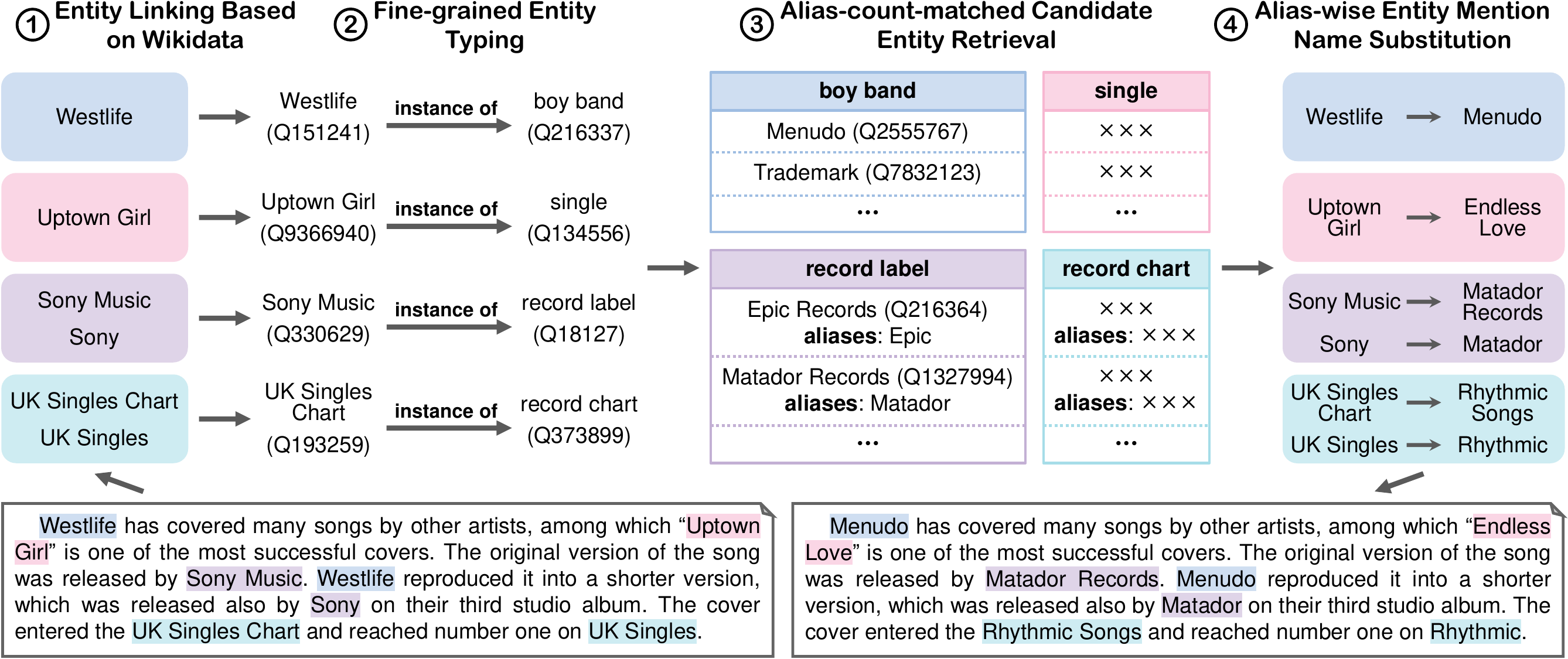}
    \caption{The proposed pipeline for generating documents with changed entity names.}
    \label{fig:pipeline}
\end{figure*}

\paragraph{Robustness and Entity-Related Robustness in NLP.} Despite achieving great progress with large pre-trained language models in various tasks, modern NLP models are still brittle to out-of-domain data \citep{hendrycks-etal-2020-pretrained}, adversarial attacks \citep{mccoy-etal-2019-right} or small perturbation to the input \citep{ebrahimi-etal-2018-hotflip}. Consequently, there has been a growing effort to explore robustness issues in NLP, such as building robustness evaluation benchmarks and proposing robustness enhancement strategies \citep{wang-etal-2022-measure, liu2024an, liu2024a, hu2024evaluating}. One branch of works focus on entity-related robustness of NLP models. By introducing various types of perturbation to entity (names), previous works audit or improve model robustness on different tasks like named entity recognition \citep{lin-etal-2021-rockner}, machine reading comprehension \citep{yan-etal-2022-robustness} and dialogue state tracking \citep{cho-etal-2022-know}. \citet{wang-etal-2023-fragile} analyse the behavior of relation extraction models under entity replacements. However, they focus on the task of sentence-level relation extraction and only consider person and organization entities.

\paragraph{Robustness of DocRE Models.} Compared with other NLP areas, research on robustness in DocRE is relatively scarce. \citet{xu-etal-2022-document} observe that DocRE models may err when non-evidence sentences of a document are removed and propose a sentence focusing loss to improve the robustness. \citet{chen-etal-2023-models} reveal the poor robustness of DocRE models to word-level attacks such as synonym substitution. A few recent works also construct entity-level attacks to investigate the robustness of DocRE models \citep{li-etal-2023-rethinking, chen-etal-2023-models}. However, all these attacks are not natural or adversarial, as they either disrupt entity structures (e.g., random entity mention drop) or alter entity types (e.g., random out-of-distribution entity substitution from a very limited source), rendering partial relation labels no longer valid. In contrast, we propose a principled pipeline to generate entity-renamed documents with labels preserved, and systematically evaluate and improve the robustness of DocRE models to entity name variations.
\section{Problem Formulation}

Given a document $D$ which contains a set of entities $\mathcal{E}=\{ e_{i} \} _{i=1}^{N_{e}} $, the task of document-level relation extraction is to predict the set of all possible relations between each entity pair $(e_{h},e_{t})\in \{ (e_{i},e_{j}) \mid i,j=1, \dots, N_{e};i\ne j \}$ from a pre-defined relation type set $\mathcal{R}$. The subscripts of $e_{h}$ and $e_{t}$ refer to the head and tail entity in an entity pair. An entity $e_{i}$ can occur multiple times in the document via $N_{e_{i}}$ mentions $\mathcal{M}_{e_{i}}=\{ m_{j}^{i} \} _{j=1}^{N_{e_{i}}} $, where the mention $m_{j}^{i}$ refers to the token span of $e_{i}$'s $j$-th occurrence in the document.
\section{Benchmark Construction}

In this section, we elaborate on the process of constructing benchmarks for evaluating the robustness of DocRE models to entity name variations. We first propose a general pipeline to generate documents with changed entity names, then apply the pipeline to DocRED and Re-DocRED to create the Env-DocRED and Env-Re-DocRED benchmarks.

\subsection{Construction Pipeline}

As shown in Figure~\ref{fig:pipeline}, our proposed pipeline consists of the following four steps.

\paragraph{Step 1: Entity Linking Based on Wikidata.} Given a document, we first link each entity in the document to an item in Wikidata. Each item in Wikidata has a label and any number of aliases, and is uniquely identified by a number starting with ``Q''. For example, we link the entity \texttt{Westlife} to item \texttt{Westlife(Q151241)} in Wikidata. Depending on the dataset at hand, we can perform entity linking using Wikidata Search API, off-the-shelf tools or methods specifically optimized for the datasets.

\paragraph{Step 2: Fine-grained Entity Typing.} Next we query the value of \texttt{Instance} \texttt{Of} property (numbered as \texttt{P31} in Wikidata) for each linked item on Wikidata, to obtain the fine-grained type of each entity, like \texttt{boy} \texttt{band(Q216337)} for \texttt{Westlife(Q151241)} in Figure~\ref{fig:pipeline}.

\paragraph{Step 3: Alias-count-matched Candidate Entity Retrieval.} Based on the fine-grained type of each entity, we further retrieve additional Wikidata items with the same entity type as candidates by executing a reverse query of Step 2. Note that we only retain those items whose number of aliases (plus label) is greater than or equal to the number of aliases of the original entity in the document. For example, since the entity \texttt{Sony} \texttt{Music} is mentioned under two different names in the document, we only take the retrieved items of \texttt{record} \texttt{label} with at least one Wikidata alias.

\paragraph{Step 4: Alias-wise Entity Mention Name Substitution.} Finally, for each entity in the document, we randomly select an item from its candidate set and use this item to perform alias-wise entity mention name substitution, i.e., substitute a distinct name of the item for each alias of the original entity, like \texttt{Sony} \texttt{Music} $\Rightarrow$ \texttt{Matador} \texttt{Records} and \texttt{Sony} $\Rightarrow$ \texttt{Matador} in Figure~\ref{fig:pipeline}.
\subsection{Env-DocRED and Env-Re-DocRED Benchmarks}

\begin{table}[t]
\centering
\resizebox{\linewidth}{!}{
\setlength{\tabcolsep}{12pt}
\begin{tabular}{lcccc}
\toprule
\multirow{2}{*}{\textbf{Type}} & \multicolumn{2}{c}{\textbf{DocRED}} & \multicolumn{2}{c}{\textbf{Re-DocRED}} \\
\cmidrule(lr){2-3} \cmidrule(lr){4-5}
& Dev & Test & Dev & Test \\
\midrule
PER & 98.52\% & 95.95\% & 98.87\% & 98.18\% \\
ORG & 95.23\% & 94.63\% & 95.07\% & 95.53\% \\
LOC & 92.88\% & 92.23\% & 93.67\% & 92.16\% \\
TIME & 99.07\% & 98.90\% & 99.51\% & 98.63\% \\
NUM & 99.67\% & 98.57\% & 99.66\% & 99.68\% \\
MISC & 78.29\% & 77.16\% & 78.40\% & 78.14\% \\
\midrule
Total & 93.39\% & 92.53\% & 93.75\% & 93.07\% \\
\bottomrule
\end{tabular}
}
\caption{Entity name substitution rates on the development and test sets of DocRED and Re-DocRED.}
\label{tab:entity_substitution_rate}
\end{table}

With the proposed pipeline, we further construct the robustness evaluation benchmarks based on existing datasets, which we choose DocRED \citep{yao-etal-2019-docred} and Re-DocRED \citep{tan-etal-2022-revisiting} in this work. DocRED is one of the largest and most popular public datasets for DocRE, which is collected from English Wikipedia documents. DocRED has 96 pre-defined relation types, along with 3053/1000/1000 documents for training/development/test. Each document in DocRED has 19.5 entities and 12.5 relation triples on average. Re-DocRED is a revised version of DocRED, resolving the missing relation issue in DocRED. The 3053 revised training documents contain 28.1 triples on average and 1000 revised development documents (split into 500/500 development/test documents) have 34.7 triples on average.

We iterate over the development and test sets of DocRED and Re-DocRED and apply the pipeline five times on each document with different random seeds. We name the two newly constructed benchmarks Env-DocRED and Env-Re-DocRED, with the former having 3053/5000/5000 and the latter having 3053/2500/2500 documents for training/development/test. We adopt the entity linking method and results of \citet{genest-etal-2023-linked} in Step 1, which has a high quality benefited from its specific design for DocRED and Re-DocRED. Besides, since all entities of NUM and TIME type in (Re-)DocRED can not be linked to Wikidata, we take a rule-based substitution method to produce novel names for number and time. Although a small portion of entities remain unlinked, statistics (Table~\ref{tab:entity_substitution_rate}) show that we have altered the name of over 92\% entities in original datasets.
\section{Robustness Evaluation and Analysis}

In this section, utilizing the constructed benchmarks, we conduct a comprehensive evaluation and analysis on the robustness of existing DocRE models to entity name variations.

\subsection{Selected Models and Evaluation Metrics}

\begin{figure*}[t]
    \centering
    \includegraphics[width=\linewidth]{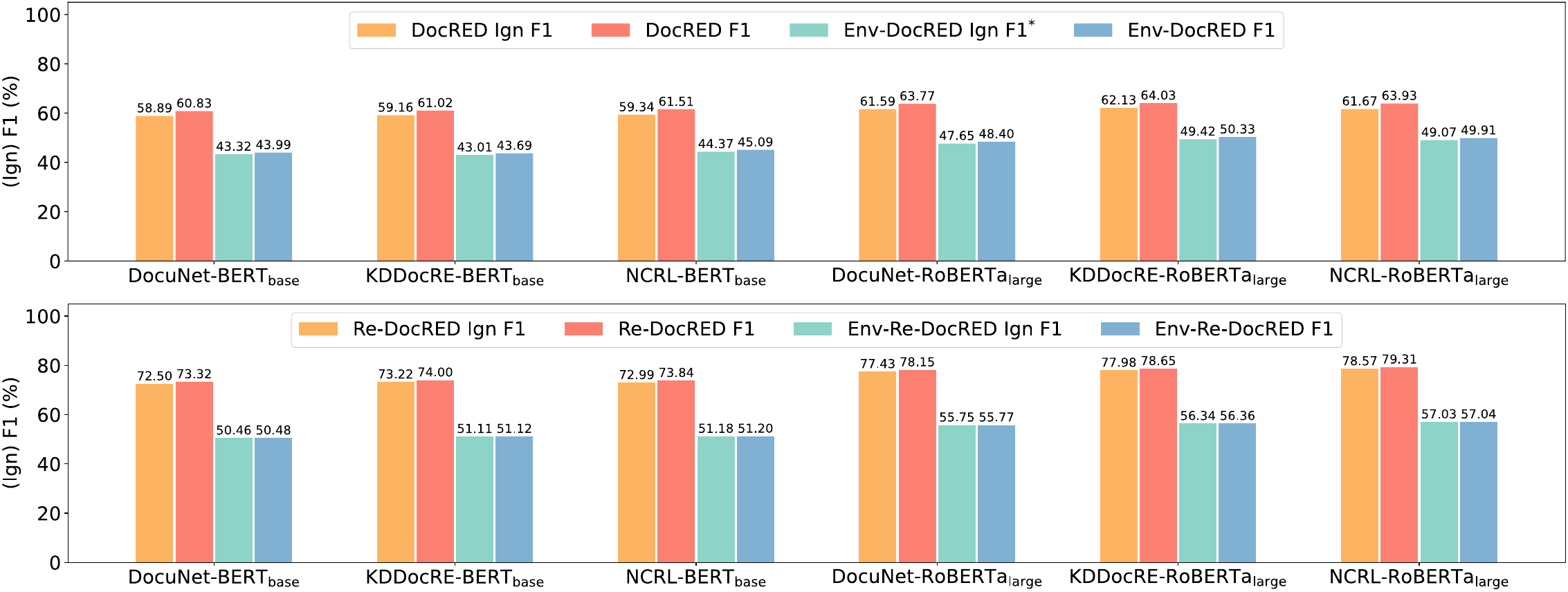}
    \caption{Evaluation results on the test sets of four benchmarks. Since the test set of DocRED is unpublished, the Ign F1 results on Env-DocRED are not accurate and marked with ``*'', same applies to Table~\ref{tab:main_test_results}.}
    \label{fig:evaluation_test_results}
\end{figure*}

We choose three public-available DocRE models which are representative for their strong performance and high popularity. \textbf{DocuNet} \citep{zhang-etal-2021-docunet} formulates DocRE as a semantic segmentation task and captures both local context information and global interdependency among triples for extraction. \textbf{KDDocRE} \citep{tan-etal-2022-document} uses an axial attention module for two-hop relation reasoning and an adaptive focal loss to address the class imbalance problem. \textbf{NCRL} \citep{zhou-lee-2022-ncrl} shares the same model with a strong DocRE baseline ATLOP \citep{zhou-etal-2021-atlop} but improves upon the learning of none class. We use Ign F1 and F1 scores as the evaluation metrics, where Ign F1 measures the F1 excluding those relational facts shared by the training and development/test sets. For each model, we all experiment with BERT$_{\mathrm{base}}$ \citep{devlin-etal-2019-bert} and RoBERTa$_{\mathrm{large}}$ \citep{liu2019roberta} encoder, leading to six submodels. We reimplement all models with their official codes and report the the mean and standard deviation results by five trials with different random seeds. Since the test set of DocRED is released by Codalab, we report the official test score of the best checkpoint on development set.
\subsection{Main Evaluation Results}

We present the evaluation results on the test sets of four benchmarks in Figure~\ref{fig:evaluation_test_results}. We can observe that all DocRE models have a significant performance fluctuation on Env-DocRED and Env-Re-DocRED, with the relative F1 drop ranging from 21\% \textasciitilde 31\%, revealing the insufficient robustness to entity name variations. Model-wise, we find that the three selected DocRE models show similar relative decline in performance, with none being significantly more robust than others. Encoder-wise, we find that RoBERTa$_{\mathrm{large}}$ with higher performance also leads to better robustness than BERT$_{\mathrm{base}}$. Dataset-wise, somewhat surprisingly, the relative decrease in F1 is even larger on Env-Re-DocRED than Env-DocRED. This suggests that despite Re-DocRED providing more complete relation labels, DocRE models still fail to gain benefits in robustness.
\subsection{Further Analysis}
\label{sec:further_analysis}

In order to gain more in-depth insights, we conduct further analysis by answering the following questions.

\subsubsection*{Q1: What is the performance bottleneck of DocRE models under entity name variations?}

Given that the entity name variations lead to a drop in performance, a natural question is whether the model generates more false positive or false negative predictions. To better understand the performance bottleneck of DocRE models, we compare the changes in precision and recall of three models with BERT$_{\mathrm{base}}$ encoder. As shown in Table~\ref{tab:precision_and_recall_analysis}, the recall across models decreases significantly, while the precision changes little and even increases on Env-DocRED. This indicates that false negative predictions dominate the poor robustness to entity name variations.

\begin{table}[t]
\centering
\resizebox{\linewidth}{!}{
\begin{tabular}{l|cccc|cccc}
\toprule
\multirow{2}{*}{\textbf{Model}} & \multicolumn{2}{c}{\textbf{DocRED}} & \multicolumn{2}{c|}{\textbf{Env-DocRED}} & \multicolumn{2}{c}{\textbf{Re-DocRED}} & \multicolumn{2}{c}{\textbf{Env-Re-DocRED}} \\
\cmidrule{2-9}
& P & R & P & R & P & R & P & R \\
\midrule
DocuNet & 62.88 & 58.67 & 64.56 & 33.23 & 84.21 & 64.93 & 82.05 & 36.45 \\
KDDocRE & 63.95 & 58.76 & 64.27 & 33.61 & 85.04 & 65.51 & 81.50 & 37.24 \\
NCRL & 63.62 & 59.08 & 65.69 & 34.50 & 84.64 & 65.50 & 81.53 & 37.32 \\
\bottomrule
\end{tabular}
}
\caption{Precision and recall results on the development sets of (Env-)DocRED and test sets of (Env-)Re-DocRED, same choices apply to Table~\ref{tab:intra_and_inter_analysis}, Figure~\ref{fig:f1_of_docs_with_different_entity_numbers} and Table~\ref{tab:ablation_study}.}
\label{tab:precision_and_recall_analysis}
\end{table}

\subsubsection*{Q2: Do models show poorer robustness when predicting inter-sentence relations?}

Since a major feature of DocRE is to extract the complex cross-sentence relations, we further analyse models' robustness in predicting intra-sentence and inter-sentence relations. We report the Intra F1 and Inter F1 of three BERT$_{\mathrm{base}}$ encoded DocRE models in Table~\ref{tab:intra_and_inter_analysis}, which respectively evaluate on the entity pairs with and without mentions in same sentence. We can observe that on both Env-DocRED and Env-Re-DocRED, the relative F1 drop for inter-sentence relations is approximately twice that of intra-sentence relations, which indicates that existing DocRE models show poorer robustness to entity name variations when predicting inter-sentence relations.

\begin{table}[t]
\centering
\resizebox{\linewidth}{!}{
\begin{tabular}{l|cccc|cccc}
\toprule
\multirow{2}{*}{\textbf{Model}} & \multicolumn{2}{c}{\textbf{DocRED}} & \multicolumn{2}{c|}{\textbf{Env-DocRED}} & \multicolumn{2}{c}{\textbf{Re-DocRED}} & \multicolumn{2}{c}{\textbf{Env-Re-DocRED}} \\
\cmidrule{2-9}
& Intra & Inter & Intra & Inter & Intra & Inter & Intra & Inter \\
\midrule
DocuNet & 66.99 & 53.11 & 52.76 & 31.34 & 76.05 & 70.92 & 58.75 & 42.27 \\
KDDocRE & 67.33 & 54.03 & 53.12 & 31.64 & 76.89 & 71.40 & 59.48 & 42.81 \\
NCRL & 67.47 & 53.84 & 54.20 & 32.58 & 76.44 & 71.57 & 59.86 & 42.51 \\
\bottomrule
\end{tabular}
}
\caption{Intra and Inter F1 results on four benchmarks.}
\label{tab:intra_and_inter_analysis}
\end{table}

\subsubsection*{Q3: How does the model robustness vary with the number of entities in the document?}

\begin{figure}[t]
    \centering
    \includegraphics[width=\linewidth]{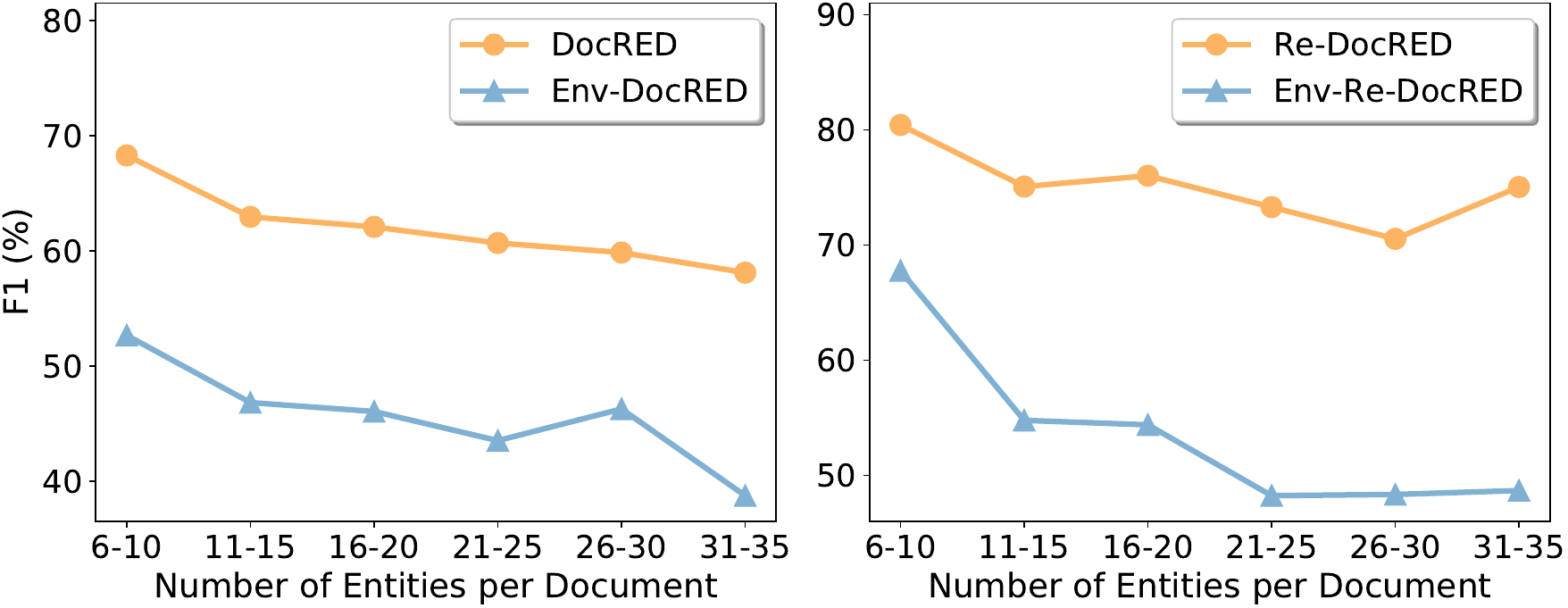}
    \caption{F1 score of NCRL-BERT$_{\mathrm{base}}$ on documents with different number of entities.}
    \label{fig:f1_of_docs_with_different_entity_numbers}
\end{figure}

We also investigate the robustness of DocRE models on documents with varying number of entities. This aids in better extrapolating our findings to longer documents, which often contain more entities. We divide the documents into different groups by the number of entities and evaluate the performance on each group. We showcase the results of NCRL-BERT$_{\mathrm{base}}$ in Figure~\ref{fig:f1_of_docs_with_different_entity_numbers}. As the number of entities increases, the absolute performance drop under entity name variations gets larger, especially on Env-Re-DocRED. The slopes of the linear fits on DocRED, Env-DocRED, Re-DocRED, Env-Re-DocRED are -0.35, -0.42, -0.24 and -0.69 respectively. Note that the performance itself also shows a decreasing trend when encountering more entities, thus the relative performance drop should be more significant. This suggests that the model may be more brittle as the number of entities increases.

\subsubsection*{Q4: How can we disentangle the reasons for the performance drop?}

\begin{table}[t]
\centering
\resizebox{\linewidth}{!}{
\begin{tabular}{l|cccccc|cccccc}
\toprule
\multirow{2}{*}{\textbf{Type}} & \multicolumn{3}{c}{\textbf{DocRED}} & \multicolumn{3}{c|}{\textbf{Env-DocRED}} & \multicolumn{3}{c}{\textbf{Re-DocRED}} & \multicolumn{3}{c}{\textbf{Env-Re-DocRED}} \\
\cmidrule{2-13}
& Q1 & Q2 & Q3 & Q1 & Q2 & Q3 & Q1 & Q2 & Q3 & Q1 & Q2 & Q3 \\
\midrule
PER & 32 & 68 & 161 & 7 & 11 & 19 & 33 & 65 & 155 & 8 & 12 & 19 \\
ORG & 27 & 104 & 587 & 7 & 12 & 27 & 34 & 125 & 685 & 7 & 13 & 28 \\
LOC & 27 & 128 & 1240 & 8 & 20 & 90 & 29 & 148 & 1704 & 8 & 21 & 108 \\
MISC & 17 & 37 & 141 & 7 & 12 & 23 & 18 & 42 & 171 & 7 & 12 & 23 \\
\midrule
Total & 25 & 73 & 309 & 7 & 13 & 29 & 27 & 81 & 393 & 7 & 13 & 32 \\
\bottomrule
\end{tabular}
}
\caption{The upper quartile (Q3), median (Q2) and lower quartile (Q1) of entity popularities of four benchmarks' test sets (only calculating entities with name changed, same applies to Table~\ref{tab:proportion_of_seen_test_entity_mentions}).}
\label{tab:popularity_of_test_entities}
\end{table}

\citet{yan-etal-2022-robustness} pointed out that the information associated with the entity name that can be leveraged by the model includes both entity knowledge and name clues. The former refers to the world knowledge associated with the entity like ``Westlife is a famous boy band'', which mainly comes from pre-training. The latter refer to the statistical clues associated with the name's surface form like ``Westlife always appears with the performer relation in training set'', which mainly come from fine-tuning. The perturbations to entity names may break these two types of information.

We adopt two measurements to better understand the information loss. We calculate the popularity of entities \citep{huang-etal-2022-recommend}, i.e., how many times the linked item of the entity appears in a relation instance in Wikidata, in each benchmark's test set to roughly quantify the entity knowledge. As shown in Table~\ref{tab:popularity_of_test_entities}, the popularity of entities in two new benchmarks drops significantly. For name clues, we calculate the percentage of entity mentions that appear in training sets for each benchmark's test set. As shown in Table~\ref{tab:proportion_of_seen_test_entity_mentions}, the proportion also has a noticeable drop in two novel benchmarks.

\begin{table}[t]
\centering
\resizebox{\linewidth}{!}{
\begin{tabular}{l@{\hspace{12pt}}|cc|cc}
\toprule
\textbf{Type} & \textbf{DocRED} & \textbf{Env-DocRED} & \textbf{Re-DocRED} & \textbf{Env-Re-DocRED} \\
\midrule
PER & 12.33\% & 1.90\% & 12.72\% & 2.23\% \\
ORG & 25.35\% & 3.47\% & 28.21\% & 3.14\% \\
LOC & 32.85\% & 8.25\% & 37.69\% & 10.77\% \\
TIME & 34.02\% & 16.62\% & 41.62\% & 20.82\% \\
NUM & 34.74\% & 12.01\% & 41.78\% & 16.86\% \\
MISC & 18.11\% & 3.04\% & 19.71\% & 3.11\% \\
\midrule
Total & 23.47\% & 5.28\% & 27.34\% & 6.89\% \\
\bottomrule
\end{tabular}
}
\caption{The proportion of entity mentions that appear in training sets of four benchmarks' test sets.}
\label{tab:proportion_of_seen_test_entity_mentions}
\end{table}

\subsubsection*{Q5: How robust is in-context learning of LLMs under entity name variations?}

\begin{table}[t]
\centering
\resizebox{\linewidth}{!}{
\setlength{\tabcolsep}{8pt}
\begin{tabular}{lcccc}
\toprule
\multirow{2}{*}{\textbf{Model}} & \multicolumn{2}{c}{\textbf{Re-DocRED}} & \multicolumn{2}{c}{\textbf{Env-Re-DocRED}} \\
\cmidrule(lr){2-3} \cmidrule(lr){4-5}
& 1-Shot & 3-Shot & 1-Shot & 3-Shot \\
\midrule
GPT-3.5 Turbo & 13.66 & 16.00 & 10.81 & 12.98 \\
GPT-4 Turbo & 28.35 & 32.41 & 21.59 & 23.08 \\
\bottomrule
\end{tabular}
}
\caption{F1 score of in-context learned LLMs on the test sets of Re-DocRED and Env-Re-DocRED.}
\label{tab:gpt_vanilla_icl_results}
\end{table}

Recently large language models (LLMs) \citep{NEURIPS2020_1457c0d6} have achieved promising few-shot results on many tasks through in-context learning (ICL) \citep{dong2023survey}. Therefore, we also conduct an experiment to explore how robust of ICL for DocRE under entity name variations. We use gpt-3.5-turbo-0125\footnote{https://platform.openai.com/docs/models/gpt-3-5-turbo} and gpt-4-0125-preview\footnote{https://platform.openai.com/docs/models/gpt-4-and-gpt-4-turbo (Due to limited budget, the experiments with gpt-4-0125-preview only use 1/5 documents.)} due to them being the most capable LLMs currently. We experiment on both 1-Shot and 3-Shot settings, which represent providing 1 and 3 example document(s) and gold relation instances as demonstrations. We randomly select demonstration document in the training set for each test document and set the temperature parameter to 0 for least randomness. The detailed prompt template is presented in Appendix~\ref{sec:appendix_llm_prompt}. The experimental results on test sets of Re-DocRED and Env-Re-DocRED are shown in Table~\ref{tab:gpt_vanilla_icl_results}. We find that on both settings, the two in-context learned LLMs show a performance drop on Env-Re-DocRED, suggesting that the robustness issue exists not only in specialized models but also in large models.
\section{Entity Variation Robust Training}

Due to the unsatisfactory robustness of existing DocRE models to entity name variations, we further explore the method for enhanced robustness. Intuitively, we can adopt a similar approach as the proposed pipeline to perturb each training document with a group of new entity names. The derived document naturally shares the same relation labels with the original one. Also, a robust DocRE model should generate consistent representations and predictions for each corresponding entity pair in the original and perturbed documents. Based on such motivation, we propose an entity variation robust training method (EVRT) that is enhanced by data augmentation and consistency regularization.

Specifically, given a labeled entity pair $(e_{h}, e_{t})$ in a document, vanilla approaches typically train the DocRE model with a classification objective $\mathcal{L}_{clo} = \ell _{task} (e_{h}, e_{t})$, where $\ell _{task}$ denotes the loss function depending on the specific model.

Denoting the corresponding entity pair of $(e_{h}, e_{t})$ in the perturbed document as $(e_{\hat{h}}, e_{\hat{t}})$, our proposed method first incorporate the classification loss $\mathcal{L}_{clp} = \ell _{task} (e_{\hat{h}}, e_{\hat{t}})$ for $(e_{\hat{h}}, e_{\hat{t}})$ to penalize the classification errors for entity pairs in the perturbed document. Then we introduce representation consistency regularization and prediction consistency regularization to encourage the model to produce consistent representations and predicted probability distributions between $(e_{h}, e_{t})$ and $(e_{\hat{h}}, e_{\hat{t}})$. Formally, we define the representation consistency regularization loss as:
\begin{equation}
    \mathcal{L}_{rcr}=\| \boldsymbol{z}^{(h, t)}-\boldsymbol{z}^{(\hat{h}, \hat{t})} \| _{2}^{2},
\end{equation}
where $\boldsymbol{z}^{(h, t)}$ is the pair representation of $(e_{h}, e_{t})$. And we define the prediction consistency regularization loss as:
\begin{equation}
    \mathcal{L}_{pcr}=\sum_{r\in \mathcal{R}} \mathcal{D}_{SKL} ( \boldsymbol{p}_{r}^{(h,t)}, \boldsymbol{p}_{r}^{(\hat{h} ,\hat{t} )} ),
\end{equation}
where $\boldsymbol{p}_{r}^{(h,t)}=[P_{r}^{(h,t)}, 1-P_{r}^{(h,t)}]$, $P_{r}^{(h,t)}$ is the predicted probability of relation $r$ for $(e_{h}, e_{t})$, $\mathcal{D}_{SKL}$ is the symmetric KL divergence:
\begin{equation}
    \mathcal{D}_{SKL}(\boldsymbol{p}, \boldsymbol{q})=\mathcal{D}_{KL}(\boldsymbol{p} \| \boldsymbol{q} )+\mathcal{D}_{KL}(\boldsymbol{q} \| \boldsymbol{p} ),
\end{equation}
where $\mathcal{D}_{KL}$ is the vanilla KL divergence. The overall objective is defined as:
\begin{equation}
    \mathcal{L}=\mathcal{L}_{clo}+\mathcal{L}_{clp}+\alpha\mathcal{L}_{rcr}+\beta\mathcal{L}_{pcr},
\end{equation}
where $\alpha$ and $\beta$ are two hyperparameters. Note that to prevent the incorporated novel entity names for training document perturbation have overlap with those entity names for substitution when constructing the benchmarks, resulting in potential shortcuts, we isolate the new entity names introduced during benchmark construction when replacing the entities in training documents.
\section{Experiments}

\subsection{Main Results}

\begin{table*}[t]
\centering
\resizebox{0.91\linewidth}{!}{
\begin{tabular}{l|cccc|cccc}
\toprule
\multirow{2}{*}{\textbf{Model}} & \multicolumn{2}{c}{\textbf{DocRED}} & \multicolumn{2}{c|}{\textbf{Env-DocRED}} & \multicolumn{2}{c}{\textbf{Re-DocRED}} & \multicolumn{2}{c}{\textbf{Env-Re-DocRED}} \\
\cmidrule{2-9}
& Ign F1 & F1 & Ign F1$^{\ast}$ & F1 & Ign F1 & F1 & Ign F1 & F1 \\
\midrule
DocuNet-BERT$_{\mathrm{base}}$ & 58.89 & 60.83 & 43.32 & 43.99 & 72.50$_{\pm\text{0.17}}$ & 73.32$_{\pm\text{0.20}}$ & 50.46$_{\pm\text{0.44}}$ & 50.48$_{\pm\text{0.44}}$ \\
\multirow{2}{*}{\quad + \textbf{EVRT}} & 58.17 & 59.71 & 51.63 & 52.78 & 71.64$_{\pm\text{0.12}}$ & 72.44$_{\pm\text{0.19}}$ & 62.32$_{\pm\text{0.46}}$ & 62.33$_{\pm\text{0.46}}$ \\
& ($\downarrow$ 0.72) & ($\downarrow$ 1.12) & ($\uparrow$ 8.31) & ($\uparrow$ 8.79) & ($\downarrow$ 0.86) & ($\downarrow$ 0.88) & ($\uparrow$ 11.86) & ($\uparrow$ 11.85) \\
\midrule
KDDocRE-BERT$_{\mathrm{base}}$ & 59.16 & 61.02 & 43.01 & 43.69 & 73.22$_{\pm\text{0.27}}$ & 74.00$_{\pm\text{0.30}}$ & 51.11$_{\pm\text{0.58}}$ & 51.12$_{\pm\text{0.58}}$ \\
\multirow{2}{*}{\quad + \textbf{EVRT}} & 58.69 & 60.21 & 51.64 & 52.94 & 72.41$_{\pm\text{0.18}}$ & 73.25$_{\pm\text{0.15}}$ & 62.53$_{\pm\text{0.19}}$ & 62.55$_{\pm\text{0.19}}$ \\
& ($\downarrow$ 0.47) & ($\downarrow$ 0.81) & ($\uparrow$ 8.63) & ($\uparrow$ 9.25) & ($\downarrow$ 0.81) & ($\downarrow$ 0.75) & ($\uparrow$ 11.42) & ($\uparrow$ 11.43) \\
\midrule
NCRL-BERT$_{\mathrm{base}}$ & 59.34 & 61.51 & 44.37 & 45.09 & 72.99$_{\pm\text{0.28}}$ & 73.84$_{\pm\text{0.32}}$ & 51.18$_{\pm\text{0.62}}$ & 51.20$_{\pm\text{0.62}}$ \\
\multirow{2}{*}{\quad + \textbf{EVRT}} & 58.84 & 60.51 & 52.97 & 54.25 & 72.00$_{\pm\text{0.36}}$ & 72.78$_{\pm\text{0.42}}$ & 62.83$_{\pm\text{0.25}}$ & 62.84$_{\pm\text{0.25}}$ \\
& ($\downarrow$ 0.50) & ($\downarrow$ 1.00) & ($\uparrow$ 8.60) & ($\uparrow$ 9.16) & ($\downarrow$ 0.99) & ($\downarrow$ 1.06) & ($\uparrow$ 11.65) & ($\uparrow$ 11.64) \\
\midrule
\midrule
DocuNet-RoBERTa$_{\mathrm{large}}$ & 61.59 & 63.77 & 47.65 & 48.40 & 77.43$_{\pm\text{0.26}}$ & 78.15$_{\pm\text{0.25}}$ & 55.75$_{\pm\text{0.70}}$ & 55.77$_{\pm\text{0.70}}$ \\
\multirow{2}{*}{\quad + \textbf{EVRT}} & 60.48 & 62.46 & 54.32 & 55.93 & 76.07$_{\pm\text{0.14}}$ & 76.68$_{\pm\text{0.18}}$ & 67.37$_{\pm\text{0.27}}$ & 67.38$_{\pm\text{0.27}}$ \\
& ($\downarrow$ 1.11) & ($\downarrow$ 1.31) & ($\uparrow$ 6.67) & ($\uparrow$ 7.53) & ($\downarrow$ 1.36) & ($\downarrow$ 1.47) & ($\uparrow$ 11.62) & ($\uparrow$ 11.61) \\
\midrule
KDDocRE-RoBERTa$_{\mathrm{large}}$ & 62.13 & 64.03 & 49.42 & 50.33 & 77.98$_{\pm\text{0.22}}$ & 78.65$_{\pm\text{0.23}}$ & 56.34$_{\pm\text{0.61}}$ & 56.36$_{\pm\text{0.61}}$ \\
\multirow{2}{*}{\quad + \textbf{EVRT}} & 60.49 & 62.20 & 56.50 & 57.83 & 76.20$_{\pm\text{0.41}}$ & 76.82$_{\pm\text{0.43}}$ & 68.60$_{\pm\text{0.25}}$ & 68.62$_{\pm\text{0.25}}$ \\
& ($\downarrow$ 1.64) & ($\downarrow$ 1.83) & ($\uparrow$ 7.08) & ($\uparrow$ 7.50) & ($\downarrow$ 1.78) & ($\downarrow$ 1.83) & ($\uparrow$ 12.26) & ($\uparrow$ 12.26) \\
\midrule
NCRL-RoBERTa$_{\mathrm{large}}$ & 61.67 & 63.93 & 49.07 & 49.91 & 78.57$_{\pm\text{0.22}}$ & 79.31$_{\pm\text{0.26}}$ & 57.03$_{\pm\text{0.94}}$ & 57.04$_{\pm\text{0.94}}$ \\
\multirow{2}{*}{\quad + \textbf{EVRT}} & 60.28 & 62.21 & 56.29 & 57.81 & 76.78$_{\pm\text{0.19}}$ & 77.48$_{\pm\text{0.21}}$ & 68.87$_{\pm\text{0.19}}$ & 68.89$_{\pm\text{0.19}}$ \\
& ($\downarrow$ 1.39) & ($\downarrow$ 1.72) & ($\uparrow$ 7.22) & ($\uparrow$ 7.90) & ($\downarrow$ 1.79) & ($\downarrow$ 1.83) & ($\uparrow$ 11.84) & ($\uparrow$ 11.85) \\
\bottomrule
\end{tabular}
}
\caption{Main results on the test sets of four benchmarks.}
\label{tab:main_test_results}
\end{table*}

The main results on the test sets of four benchmarks are shown in Table~\ref{tab:main_test_results}. It is shown that when equipped with the proposed EVRT method, all DocRE models achieve a significant performance gain on Env-DocRED (a maximum more then 9\% absolute increase in F1) and Env-Re-DocRED (a maximum more than 12\% absolute increase in F1). Meanwhile, the performance on DocRED and Re-DocRED only shows a slight drop. All these results indicate that EVRT can effectively improve the robustness of existing DocRE models to entity name variations.
\subsection{Ablation Study}

\begin{table}[t]
\centering
\renewcommand\arraystretch{0.9}
\resizebox{\linewidth}{!}{
\begin{tabular}{c@{\hspace{16pt}}c@{\hspace{16pt}}ccccc}
\toprule
\multirow{2}{*}{$\boldsymbol{\mathcal{L}_{clp}}$} & \multirow{2}{*}{$\boldsymbol{\mathcal{L}_{rcr}}$} & \multirow{2}{*}{$\boldsymbol{\mathcal{L}_{pcr}}$} & \multicolumn{2}{c}{\textbf{Env-DocRED}} & \multicolumn{2}{c}{\textbf{Env-Re-DocRED}} \\
\cmidrule(lr){4-5} \cmidrule(lr){6-7}
& & & Ign F1 & F1 & Ign F1 & F1 \\
\midrule
$\boldsymbol{-}$ & $\boldsymbol{-}$ & $\boldsymbol{-}$ & 45.21 & 45.23 & 51.18 & 51.20 \\
\Checkmark & $\boldsymbol{-}$ & $\boldsymbol{-}$ & 52.89 & 52.91 & 62.05 & 62.06 \\
$\boldsymbol{-}$ & \Checkmark & $\boldsymbol{-}$ & 52.13 & 52.14 & 61.08 & 61.10 \\
$\boldsymbol{-}$ & $\boldsymbol{-}$ & \Checkmark & 53.36 & 53.38 & 61.83 & 61.84 \\
\Checkmark & \Checkmark & $\boldsymbol{-}$ & 52.75 & 52.77 & 62.21 & 62.22 \\
\Checkmark & $\boldsymbol{-}$ & \Checkmark & 53.79 & 53.80 & 62.41 & 62.42 \\
$\boldsymbol{-}$ & \Checkmark & \Checkmark & 53.50 & 53.52 & 62.09 & 62.11 \\
\Checkmark & \Checkmark & \Checkmark & \textbf{54.15} & \textbf{54.17} & \textbf{62.83} & \textbf{62.84} \\
\bottomrule
\end{tabular}
}
\caption{Ablation study results.}
\label{tab:ablation_study}
\end{table}

We further conduct an ablation study on Env-DocRED and Env-Re-DocRED to investigate the influence of three newly added training objectives. As shown in Table~\ref{tab:ablation_study}, only introducing one of $\mathcal{L}_{clp}$, $\mathcal{L}_{rcr}$ and $\mathcal{L}_{pcr}$ has lead to a significant performance improvement, which indicates the effectiveness of each objective. When combining these losses pairwise, the performance is further enhanced. And the best performance is achieved when simultaneously using three objectives together. We also observe that compare to $\mathcal{L}_{rcr}$, $\mathcal{L}_{clp}$ and $\mathcal{L}_{pcr}$ may play a more important role for the improvement.
\subsection{Understanding and Reasoning Capability Evaluation}

\begin{figure}[t]
    \centering
    \includegraphics[width=0.9\linewidth]{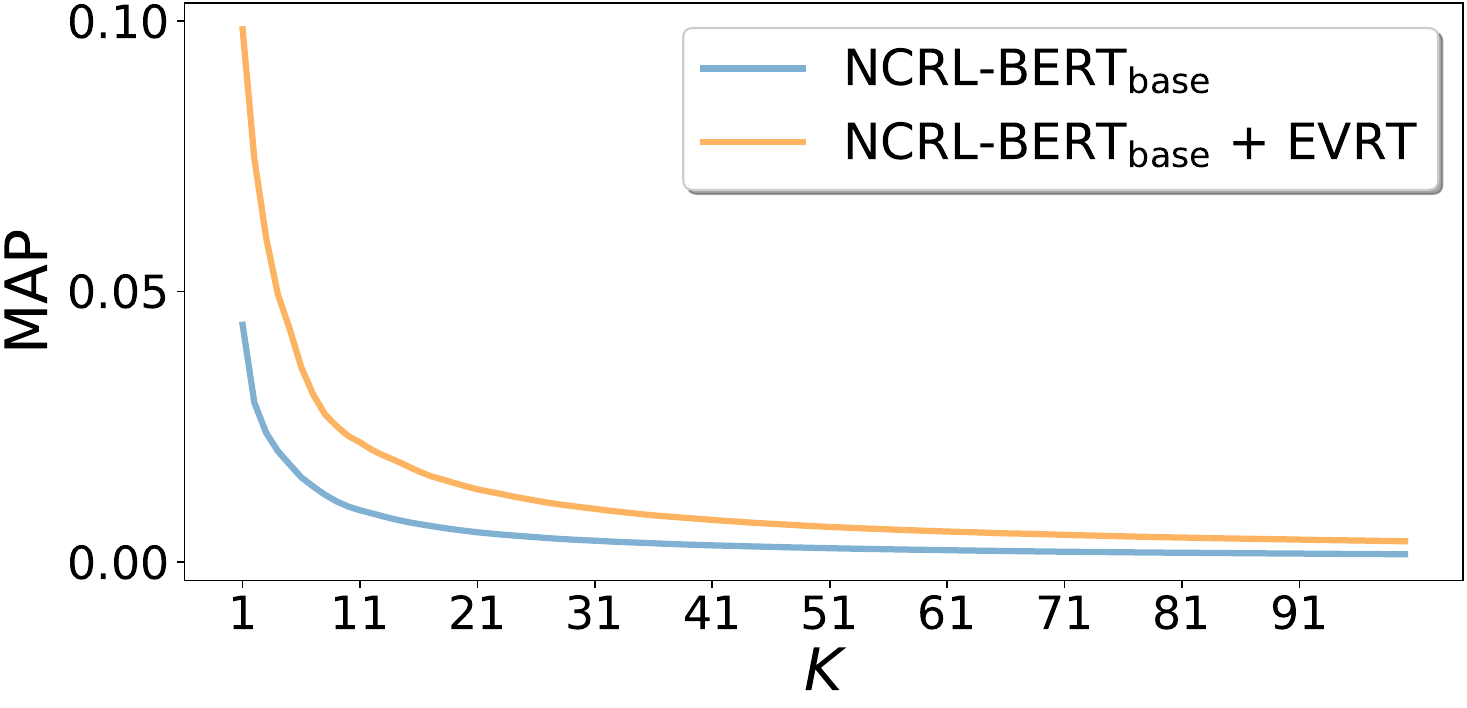}
    \caption{MAP curves of NCRL-BERT$_{\mathrm{base}}$ and NCRL-BERT$_{\mathrm{base}}$ + EVRT.}
    \label{fig:map_curve}
\end{figure}

We also take the MAP evaluation metric proposed in \citet{chen-etal-2023-models} to evaluate the understanding and reasoning capabilities of DocRE models trained with and without our EVRT method. Given top $K$ words with the highest attribution values, the formula of MAP over $T$ relational facts is:
\begin{equation}
\small
    \mathrm{MAP}(K)=\frac{1}{T} \sum_{t=1}^{T} \mathrm{AP}_{t}(K)=\frac{1}{T} \sum_{t=1}^{T} \frac{1}{K} \sum_{i=1}^{K} P_{t}(i) \cdot \mathbf{1}_{t}(i),
\end{equation}
where $\mathbf{1}_{t}(i)$ is the indicator function of the $i$-th important word for predicting the $t$-th relational fact. We select all possible values of $K$ and report the MAP curve of NCRL-BERT$_{\mathrm{base}}$ and NCRL-BERT$_{\mathrm{base}}$ + EVRT models in Figure~\ref{fig:map_curve}. It is observed that the MAP values of NCRL-BERT$_{\mathrm{base}}$ + EVRT are consistently higher than NCRL-BERT$_{\mathrm{base}}$, suggesting that the proposed EVRT method not only improves the robustness of DocRE models but also enhances their understanding and reasoning capabilities.
\subsection{Entity Variation Robust In-Context Learning}

\begin{table}[t]
\centering
\resizebox{\linewidth}{!}{
\setlength{\tabcolsep}{8pt}
\begin{tabular}{lcccc}
\toprule
\multirow{2}{*}{\textbf{Model}} & \multicolumn{2}{c}{\textbf{Re-DocRED}} & \multicolumn{2}{c}{\textbf{Env-Re-DocRED}} \\
\cmidrule(lr){2-3} \cmidrule(lr){4-5}
& 1-Shot & 3-Shot & 1-Shot & 3-Shot \\
\midrule
GPT-3.5 Turbo & 13.66 & 16.00 & 10.81 & 12.98 \\
\quad + DA & 14.67 & 16.47 & 11.59 & 13.86 \\
\quad + DA + CG & \textbf{15.14} & \textbf{17.22} & \textbf{12.44} & \textbf{14.37} \\
\midrule
GPT-4 Turbo & 28.35 & 32.41 & 21.59 & 23.08 \\
\quad + DA & 28.20 & 33.52 & 22.85 & 24.41 \\
\quad + DA + CG & \textbf{28.99} & \textbf{34.32} & \textbf{23.74} & \textbf{25.11} \\
\bottomrule
\end{tabular}
}
\caption{F1 score of entity variation robust in-context learning method for DocRE.}
\label{tab:gpt_evr_icl_results}
\end{table}

The results in Section~\ref{sec:further_analysis} indicate that utilizing in-context learning of LLMs for DocRE also shows insufficient robustness to entity name variations. A natural question is can we transfer the basic idea of EVRT to improve the robustness of in-context learning. We conduct a preliminary attempt by designing a simple entity variation robust in-context learning method, which optimizes the prompt with demonstration augmentation (DA) and consistency guidance (CG). Based on the vanilla prompt, demonstration augmentation adds an entity-renamed document for each original demonstration document. And consistency guidance further expands the prompt by explicitly explaining that ``\textit{The only difference between two documents lies in the entity names. Apart from the entities, the contextual content of the two documents is entirely the same. Therefore, the expected outputs for the two documents are also identical. When extracting relation triples from the test document, please base the extraction on the context of the document and avoid identifying the relations solely based on the information of the entities themselves.}''. As shown in Table~\ref{tab:gpt_evr_icl_results}, this simple strategy also effectively enhances the robustness of LLM-based in-context learning methods.
\section{Conclusion}

In this work, we systematically study the robustness of DocRE models to entity name variations. Our main contributions are three-fold: (1) Resource-wise, we propose a general pipeline to reasonably generate entity-renamed documents and construct two novel benchmarks, Env-DocRED and Env-Re-DocRED, for robustness evaluation. (2) Experiment-wise, we conduct comprehensive experiments on multiple DocRE models to evaluate their robustness and provide further analyses from multiple perspectives. (3) Methodology-wise, we propose entity variation robust training and in-context learning methods, effectively improving the robustness of DocRE models. We hope our work can benefit and offer insights for future research to develop more robust DocRE models.
\section{Limitations and Future Directions}
\label{sec:limitations_and_future_directions}

In this section, we analyse the limitations of our work from three perspectives and hope to provide inspiration for future works.

\paragraph{Task Setting.} Our study is grounded upon a classic setting of DocRE where the entity information including entity mention positions and coreference clusters of mentions are given beforehand. Some recent works explore the end-to-end setting of DocRE, which requires the model to jointly perform mention detection (and optionally classification), coreference resolution and relation extraction, aligning better with real-world application scenarios \citep{eberts-ulges-2021-end, giorgi-etal-2022-sequence, xu-choi-2022-modeling, zhang-etal-2023-novel}. Investigating the robustness of end-to-end DocRE approaches to entity name variations is a promising direction for future works. More importantly, since the proposed pipeline for entity name substitution does not alter entity types and coreference labels, our constructed benchmarks can be directly utilized for the study of end-to-end DocRE model robustness, rendering the two benchmarks more valuable.

\paragraph{Dataset Domain and Language.} Given that we construct the robustness evaluation benchmarks based on DocRED and Re-DocRED, which originate from English Wikipedia documents, our findings may be somewhat limited to English, generic-domain scenarios. Leveraging other well-established DocRE datasets, future works are encouraged to extend the study on entity name variation robustness of DocRE models to more domains such as news \citep{ZAPOROJETS2021102563}, biomedicine \citep{10.1093/database/baw068}, social media \citep{10.1145/3581783.3611899} and scientific publications \citep{luan-etal-2018-multi}, and more languages such as Chinese \citep{cheng-etal-2021-hacred} and Korean \citep{yang-etal-2023-histred}. As Wikidata covers a wide range of domains and languages, the proposed benchmark construction pipeline can also be applied to other datasets. For datasets that are hard to be linked to Wikidata, one may explore the possibility of adapting the pipeline with an appropriate knowledge base.

\paragraph{Methodology.} Since the proposed entity variation robust training and in-context learning frameworks generate a perturbed document with changed entity names for each training document, fine-tuning pre-trained models incurs larger memory overhead, and utilizing large language models for in-context learning entails higher time and cost expenses. Additionally, although the proposed methods significantly improve the performance of multiple models on Env-DocRED and Env-Re-DocRED, there is still a certain gap compared to DocRED and Re-DocRED. An intriguing avenue for future research is to explore more efficient and effective techniques to improve the robustness of DocRE models to entity name variations.

\bibliography{custom}

\appendix

\section{In-Context Learning Prompt Template for DocRE (1-Shot as Example)}
\label{sec:appendix_llm_prompt}

\textbf{In-context learning prompt template for DocRE (1-shot as example):}

\textit{Given a document in which all entity mentions have been marked, please identify all relation types between any two different entities based on the context of the document. The scope of target relation types for identification is limited to these 96 types (separated by semicolons): head of government; country; place of birth; place of death; father; mother; spouse; country of citizenship; continent; instance of; head of state; capital; official language; position held; child; author; member of sports team; director; screenwriter; educated at; composer; member of political party; employer; founded by; league; publisher; owned by; located in the administrative territorial entity; genre; operator; religion; contains administrative territorial entity; follows; followed by; headquarters location; cast member; producer; award received; creator; parent taxon; ethnic group; performer; manufacturer; developer; series; sister city; legislative body; basin country; located in or next to body of water; military branch; record label; production company; location; subclass of; subsidiary; part of; original language of work; platform; mouth of the watercourse; original network; member of; chairperson; country of origin; has part; residence; date of birth; date of death; inception; dissolved, abolished or demolished; publication date; start time; end time; point in time; conflict; characters; lyrics by; located on terrain feature; participant; influenced by; location of formation; parent organization; notable work; separated from; narrative location; work location; applies to jurisdiction; product or material produced; unemployment rate; territory claimed by; participant of; replaces; replaced by; capital of; languages spoken, written or signed; present in work; sibling. Entities in the document are numbered in the order of their first mention, and each entity mention is enclosed in the corresponding entity number. Before the test document, an example document and its expected output are provided. Please output the extraction results of the test document in the same format as the example, i.e., each line outputs an extracted relation triple, and the format of each triple is: <subject entity number; relation type; object entity number>. Each relation triple should be output only once.}

\textit{Example document:}

\textit{......}

\textit{All relation triples extracted from the document:}

\textit{......}

\textit{Test document:}

\textit{......}

\textit{All relation triples extracted from the document:}

\end{document}